\documentclass[journal]{IEEEtai}

\usepackage[colorlinks,urlcolor=blue,linkcolor=blue,citecolor=blue]{hyperref}
\usepackage{color,array}
\usepackage{graphicx}

\usepackage{times}
\usepackage{soul}
\usepackage{url}
\usepackage[small]{caption}
\usepackage{graphicx}
\usepackage{amsmath}
\usepackage{amsthm}
\usepackage{booktabs}
\usepackage{algorithm}
\usepackage{algorithmic}
\usepackage[switch]{lineno}
\usepackage{amsmath,amssymb,amsfonts}
\usepackage{xcolor}
\usepackage{hyperref}
\usepackage{tikz}
\usepackage{booktabs}
\usepackage{makecell}
\usepackage{xspace}

\newcommand{\xmark}{%
\tikz[scale=0.23] {
    \draw[line width=0.7,line cap=round] (0,0) to [bend left=6] (1,1);
    \draw[line width=0.7,line cap=round] (0.2,0.95) to [bend right=3] (0.8,0.05);
}}
\newcommand{\cmark}{%
\tikz[scale=0.23] {
    \draw[line width=0.7,line cap=round] (0.25,0) to [bend left=10] (1,1);
    \draw[line width=0.8,line cap=round] (0,0.35) to [bend right=1] (0.23,0);
}}

 \newcommand{\smquad}{\hspace{0.1em}} 

\usepackage{tgpagella}
\usepackage{xcolor,colortbl}
\usepackage{hyperref}
\hypersetup{
     colorlinks=true,
     linkcolor=blue,
     filecolor=magenta,      
     urlcolor=blue,
     citecolor=blue,
 }
\usepackage{balance}

\begin{document}

\title{CoSwin: Convolution Enhanced Hierarchical Shifted Window Attention For\\ Small-Scale Vision} 

\author{
    Puskal Khadka, Rodrigue Rizk, Longwei Wang, KC Santosh\\
    AI Research Lab, Department of Computer Science,\\
    University of South Dakota, Vermillion, SD, USA.\\
    puskal.khadka@coyotes.usd.edu, \{rodrigue.rizk, longwei.wang, kc.santosh\}@usd.edu
}

\maketitle
\thispagestyle{empty} 
\pagestyle{empty}

\begin{abstract}
Vision Transformers (ViTs) have achieved impressive results in computer vision by leveraging self-attention to model long-range dependencies. However, their emphasis on global context often comes at the expense of local feature extraction in small datasets, particularly due to the lack of key inductive biases such as locality and translation equivariance. To mitigate this, we propose \textbf{CoSwin}, a novel feature-fusion architecture that augments the hierarchical shifted window attention with localized convolutional feature learning. Specifically, CoSwin integrates a learnable local feature enhancement module into each attention block, enabling the model to simultaneously capture fine-grained spatial details and global semantic structure. We evaluate CoSwin on multiple image classification benchmarks including CIFAR-10, CIFAR-100, MNIST, SVHN, and Tiny ImageNet. Our experimental results show consistent performance gains over state-of-the-art convolutional and transformer-based models. Notably, CoSwin achieves improvements of \textbf{2.17\%} on CIFAR-10, \textbf{4.92\%} on CIFAR-100, \textbf{0.10\%} on MNIST, \textbf{0.26\%} on SVHN, and \textbf{4.47\%} on Tiny ImageNet over the baseline Swin Transformer. These improvements underscore the effectiveness of local-global feature fusion in enhancing the generalization and robustness of transformers for small-scale vision. Code and pretrained weights available at \url{https://github.com/puskal-khadka/coswin}.
\end{abstract}

\begin{IEEEImpStatement}
CoSwin advances the design of Vision Transformers by embedding local feature learning into shifted window attention, thereby overcoming the absence of key inductive biases such as locality and translation equivariance. The resulting architecture not only improves classification accuracy across multiple benchmark datasets but also demonstrates stronger generalization on small-scale and limited datasets, where traditional transformer models have struggled. By providing a scalable and practical solution for robust image understanding, the proposed methodology provides a practical and scalable solution for robust image understanding, especially in small-scale vision datasets.
\end{IEEEImpStatement}


\section{Introduction}
The field of deep learning for computer vision has traditionally been dominated by Convolutional Neural Networks (CNNs)~\cite{cite_cnn}, which have proven highly effective for tasks such as image classification, object detection~\cite{cite_object_detection}, and semantic segmentation~\cite{cite_segmentation}. Numerous CNN architectures~\cite{cite_resnet,cite_densenet,cite_xception,cite_resnetxt} have become standard choices for image analysis due to their powerful ability to capture detailed local features and their strong inductive biases. These properties make CNNs inherently well-suited for learning robust representations from structured grid-like data such as images.

Meanwhile, the development of transformer architectures~\cite{cite_transformer} has brought revolutionary advancements in natural language processing~\cite{cite_bert,cite_language_model_gpt3,cite_roberta}. Their success, primarily attributed to the self-attention mechanism capable of modeling long-range dependencies, has inspired researchers to adapt transformers for vision tasks. Early efforts~\cite{cite_attn_conv,cite_attn_img_recog,cite_bottleneck_transformer,cite_pyramid_vit} explored integrating attention-based mechanisms for computer vision tasks to address some of the limitations of CNN in capturing global context.

Among these efforts, Vision Transformer (ViT)~\cite{cite_vit} and Swin Transformer~\cite{cite_swin} have emerged as prominent architectures, achieving state-of-the-art performance on various benchmarks. These models treat images as sequences of patches and apply self-attention to model interactions between patches which allows a comprehensive understanding of long-range dependencies~\cite{cite_vit_survey}. However, despite their success, pure transformer-based vision models face critical challenges. The self-attention mechanism, while excellent at global context modeling, often struggles to focus on localized patterns of small datasets critical for precise image recognition. Furthermore, vision transformers inherently lack key inductive biases such as translation equivariance and locality, which are naturally embedded in CNNs. This absence of structural priors makes transformers heavily data-dependent and less effective when trained on limited datasets~\cite{cite_vit}. As a result, vision transformers may require larger datasets, or additional regularization techniques to generalize effectively.

To mitigate these challenges, we propose a deep learning model named \textbf{CoSwin}, which integrates weighted local feature extraction into the shifted window attention mechanism. By fusing weighted convolutional feature maps with self-attention scores, CoSwin is designed to effectively capture both local spatial patterns and global semantic dependencies even in small datasets. Specifically, our model enhances the traditional attention mechanism by incorporating a learnable local feature enhancement module, thereby adding crucial inductive biases into the model while preserving its ability to perform long-range modeling.

Our contributions can be summarized as follows:
\begin{itemize}
    \item We propose CoSwin, a novel architecture that integrates weighted local feature enhancement into the shifted window attention framework. This architecture improves both local details and global context modeling for small scale vision datasets. 
    \item We implement a patch conversion technique that reshapes linear patch embeddings into an imaginary image format. It enables effective convolution operations over patch-based transformer inputs without disrupting the attention structure.
    \item Following our experiments on small scale vision datasets such as CIFAR-10/100~\cite{cite_cifar}, MNIST~\cite{cite_mnist}, SVHN~\cite{cite_svhn}, and Tiny ImageNet~\cite{cite_mnist}, we demonstrate that CoSwin achieves significant improvements over the SOTA baselines, particularly showcasing enhanced robustness and data efficiency by incorporating inductive biases into the transformer framework.
\end{itemize}

\section{Related Work}
\subsection{Convolutional Neural Networks:}
CNNs have long served as the foundation for modern image classification. They excel at learning hierarchical feature representations by progressively extracting local patterns such as edges, textures, and shapes. Early architectures such as AlexNet~\cite{cite_alexnet} and VGG~\cite{cite_vgg} significantly advanced the field by demonstrating the effectiveness of deep convolutional layers in improving classification accuracy. Subsequent developments like ResNet~\cite{cite_resnet} introduced skip connections to mitigate the vanishing gradient problem and enable the successful training of much deeper networks.

Despite their widespread success, CNNs face limitations in modeling global context due to their inherently localized receptive fields affecting adversely in small training datasets. Although techniques such as dilated convolutions~\cite{cite_dilated_conv} have been proposed to expand receptive fields, CNNs still struggle to capture long-range dependencies effectively. 

\subsection{Vision Transformers:}
ViTs~\cite{cite_vit} and their variants~\cite{cite_multi_vit,cite_mobile_vit,cite_flexi_vit,cite_t2t} represent a paradigm shift from traditional convolution-based approaches by treating images as sequences of fixed-size, non-overlapping patches. These patches are linearly embedded into feature vectors and then processed using self-attention mechanisms across transformer layers. Unlike CNNs, which hierarchically aggregate features through localized convolutional operations~\cite{cite_cnn,cite_gradient_doc}, ViTs model pairwise relationships across all patches to capture long-range dependencies~\cite{cite_vit}.

However, a major challenge with standard Vision Transformers is that they lack strong inductive biases, which makes them less effective on small datasets~\cite{cite_deit}. To tackle this, several studies have proposed improvements like DeiT~\cite{cite_deit} used knowledge distillation to make training more data-efficient. Chhabra et al.~\cite{cite_patchrot} proposed a self-supervised approach to address the need for labeling large datasets in ViTs. Similarly, for small datasets, Lee et al.~\cite{cite_lee} introduced shifted patch tokenization and locality-aware attention to improve ViT training. Gani et al.~\cite{cite_gani} proposed a self-supervised weight learning strategy, while Liu et al.~\cite{cite_liu} introduced a self-supervised auxiliary task to regularize ViT training. While these methods help, ViTs still face challenges when training from scratch on limited datasets.

\subsection{Swin Transformer:}
The Swin Transformer~\cite{cite_swin} introduced a significant innovation in vision transformer architectures by proposing a hierarchical design based on non-overlapping shifted windows. Instead of computing global self-attention across the entire image, Swin Transformer restricts attention computation within local windows and progressively shifts these windows across layers. This hierarchical windowing strategy reduces computational complexity while preserving the ability to model image contexts over successive stages.

However, while Swin's window-based attention mechanism captures localized patterns to some extent, it still fundamentally relies on attention operations without explicitly embedding strong inductive biases such as translation equivariance~\cite{cite_swin}. Consequently, the Swin Transformer can still struggle to capture fine-grained local features, particularly in scenarios involving less training data and low-quality images.\\
Our proposed model builds upon the architectural foundations of vision transformers but introduces a critical enhancement: fusing weighted local feature learning into the shifted window attention mechanism. By integrating a learnable local feature enhancement module alongside attention operations, we aim to leverage the inherent advantages of convolutions in capturing fine spatial details while retaining the transformer’s strength in modeling long-range global dependencies. This balanced approach enables our model to effectively focus on both local and global information in parallel, leading to improved efficiency and generalization in image classification tasks regardless of small-size datasets.

\begin{figure*}[t!]
\centering
\includegraphics[width=0.95\textwidth]{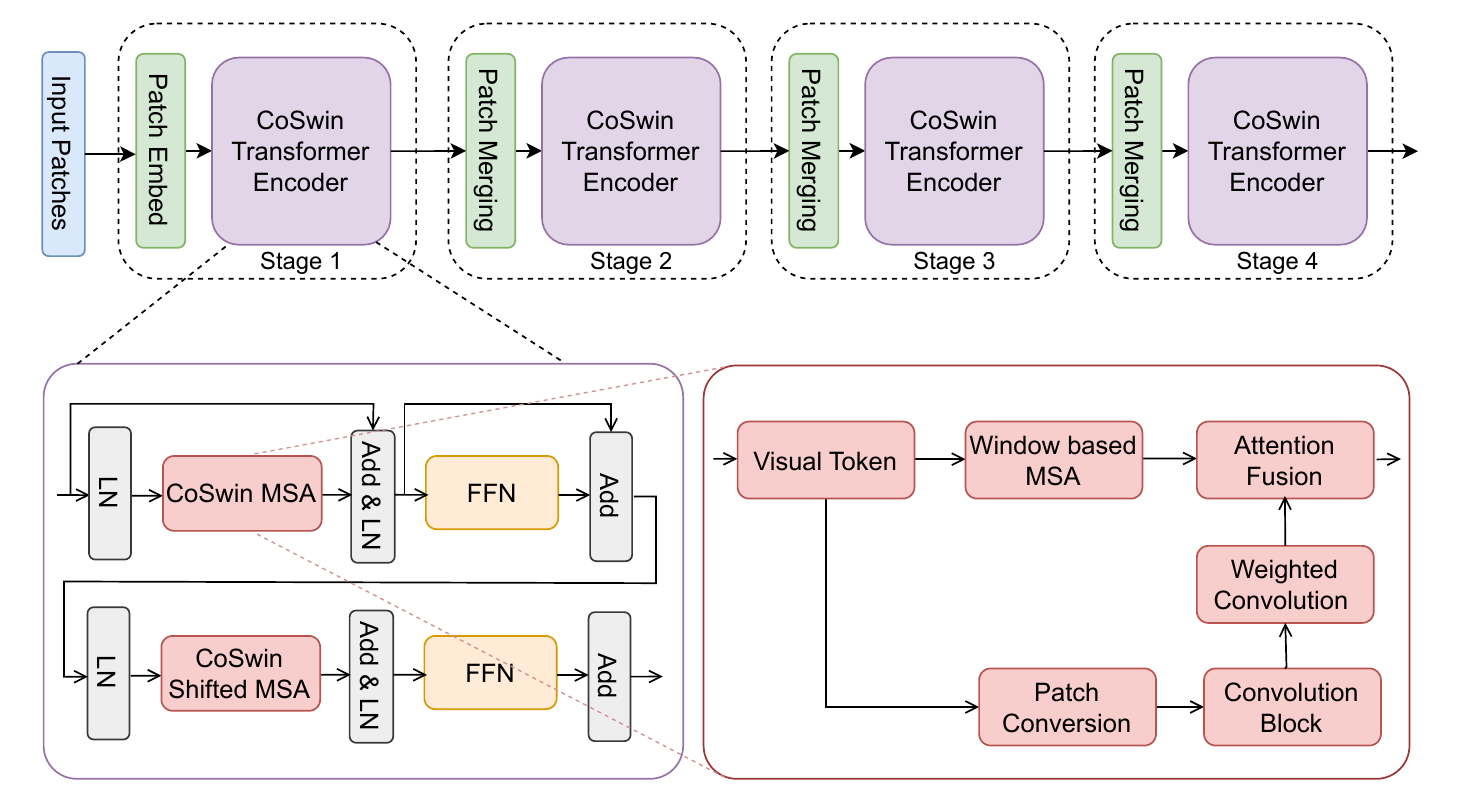}
\caption{
(Top) Proposed CoSwin architecture. The input image is split into non-overlapping patches, embedded linearly, and then processed by CoSwin encoder blocks. (Bottom Left) Two successive CoSwin encoder blocks. (Bottom Right) CoSwin-MSA architecture, where attention scores are refined using weighted convolution through multiple layers.}
\label{fig:coswin}
\end{figure*}

\begin{figure*}[t!]
\centering
\includegraphics[width=1\textwidth]{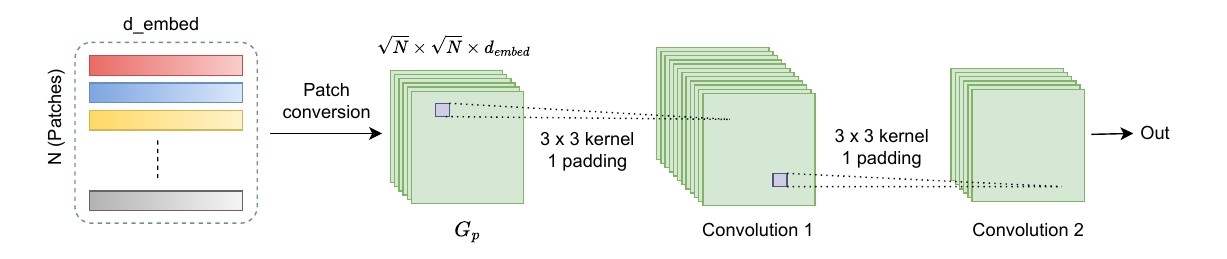}
\caption{Architecture of Convolution Blocks in Local Feature Enhancement Module.}
\label{fig:conv_arch}
\end{figure*}

Our proposed model builds upon the architectural foundations of vision transformers but introduces a critical enhancement: fusing weighted local feature learning into the shifted window attention mechanism. By integrating a learnable local feature enhancement module alongside attention operations, we aim to leverage the inherent advantages of convolutions in capturing fine spatial details while retaining the transformer’s strength in modeling long-range global dependencies. This balanced approach enables our model to effectively focus on both local and global information, leading to improved efficiency and generalization in image classification tasks across diverse datasets.

\section{Proposed Method}
In this section, we present the detailed architecture of our proposed model, \textbf{CoSwin}, and describe its core components.

\subsection{CoSwin Architecture}
The foundation of the proposed CoSwin model follows the standard vision transformer scheme with four stages. All stages share similar architecture with uniformly decreasing spatial resolution and increasing channel numbers. Each stage consists of an embedding layer followed by a series of $n$ CoSwin Transformer blocks.

Given an image $ \mathbf{I} \in \mathbb{R}^{H \times W \times C} $,
where $H$ and $W$ is the spatial resolution of an image and $C$ represents channels, we divide image into $P\times P$ non-overlapping patches. With these patches, an image can be represented as $ \mathbf{I_p} \in \mathbb{R}^{N \times (P^2\times C)} $, where $ N = \frac{H}{P} \times \frac{W}{P}$ is the total number of visual patches. In the first stage, we use a patch embedding layer to convert each visual patch into a token represented by a $d_{embed}$ dimensional vector, resulting in an image $X$ in the tokenized format. From the second to the fourth stage, we use a patch merging layer to reduce the number of tokens, building the hierarchical feature representation. The output resolution of the patch merging layer for $(i+1)^{th}$  stage, i.e., $Merge_{i+1}$, can be expressed as
\begin{equation}
\small
  \mbox{Merge}_{i+1} = \frac{H}{2\times P_{i}} \times \frac{W}{2\times P_{i}} \times 2(d_{embed})_{i}\smquad{.}
  \label{eq:merge_stage1}
\end{equation}
Considering our architecture with a default patch size of 4, our model has
$\frac{H}{4} \times \frac{W}{4}$, $\frac{H}{8} \times\frac{W}{8}$, $\frac{H}{16} \times \frac{W}{16}$ and $\frac{H}{32} \times \frac{W}{32}$ tokens respectively, where spatial resolution is reducing gradually by twice in each stage.

The embedding layer in the first stage and the patch merging layer in the remaining stages are followed by a set of CoSwin Transformer blocks. In each block, we use the convolution fused window multi-head self-attention module (shifted and non-shifted) and a feed-forward network (FFN). This fused multi-head attention module integrates both weighted convolution and attention mechanisms, as illustrated in Figure~\ref{fig:coswin}. 
Both attention and feed-forward networks are preceded by layer normalization (LN)~\cite{cite_layer_norm} and followed by residual connection to improve training.

\subsection{Shifted Window Attention with Learnable Local Features}
\label{sec:coswin_msa}

The core innovation of CoSwin is its feature-fused attention mechanism, which is designed to jointly capture fine-grained local details and long-range global dependencies. Extending the shifted window attention framework~\cite{cite_swin}, CoSwin introduces a learnable local feature enhancement module that is adaptively fused with the multi-head attention output to embed locality into the hierarchical structure.\\

\noindent \textbf{Patch Conversion for Convolution Compatibility:}  
To enable convolutional operations within the transformer framework, we introduce a patch conversion layer that reshapes the sequence of patch tokens into a pseudo-image format. Given $N$ patch tokens with embedding dimension $d_{\text{embed}}$, where $N$ is a perfect square, the reshaped feature map is defined as:
\begin{equation}
G_p \in \mathbb{R}^{\sqrt{N} \times \sqrt{N} \times d_{\text{embed}}},
\label{eq:patch_conversion}
\end{equation}
where $\sqrt{N} \times \sqrt{N}$ corresponds to the spatial resolution and $d_{\text{embed}}$ denotes the channel dimension.\\

\noindent \textbf{Local Feature Extraction via Convolution:}  
A two-layer $3 \times 3$ convolution block is applied to $G_p$ with padding 1 to preserve spatial dimensions. The first convolution expands the number of channels by 10\%, followed by ReLU activation, and the second projects the features back to $d_{\text{embed}}$ channels:
\begin{equation}
     F_{\rm conv}= \text{Conv}_{3\times3} \left(\sigma(\text{Conv}_{3\times3}(G_{p})) \right),
\label{eq:deformable}
\end{equation}
where $\sigma(\cdot)$ denotes the ReLU activation function. To control the influence of the convolutional features relative to the attention output, we introduce a learnable scalar weight $\Gamma$:
\begin{equation}
F_{\text{weighted\_conv}} = \Gamma \times F_{\text{conv}}.
\label{eq:wt_conv}
\end{equation}
\\
\noindent \textbf{Window-Based Self-Attention and Fusion:}  
Following standard window partitioning, the tokenized input $X$ is divided into $S$ non-overlapping windows, each of size $M \times M$ patches, where S will be equal to $ \frac{H \times W}{M^2}$. Then the part of image $X$ belonging to $i^{th}$ window can be represented as
\begin{equation}
X_i \in \mathbb{R}^{M^2 \times d_{\text{embed}}}.
\end{equation}
Within each window, multi-head self-attention $MSA(X_i)$ is computed by applying self attention across k heads and concatenating their output as follows:
\begin{equation}
\text{MSA}(X_i) = \text{concat}[\text{head}_1, \ldots, \text{head}_k],
\end{equation}
where each head is defined as:
\begin{equation}
\text{head}_k = \text{Attention}(Q_k, K_k, V_k),
\end{equation}
and for a single attention head, we computed the corresponding attention value as follows:
\begin{equation}
\text{Attention}(Q, K, V) = \text{Softmax}\left(\frac{QK^\top}{\sqrt{d}} + B\right)V.
\end{equation}
Here, $Q, K, V \in \mathbb{R}^{M^2 \times d}$ are the query, key, and value matrices, $d$ is the head dimension, and $B$ encodes the relative positional bias.

The multi-head attention scores from all windows are then merged and fused with the weighted convolution score from Eq.~\ref{eq:wt_conv} to derive the convolution fused window-based attention score, i.e., CoSwin-MSA(X), which can be mathematically computed as follows:
\begin{equation}
\text{CoSwin-MSA}(X) = \left( \bigcup_{i=1}^{S} \text{MSA}(X_i) \right) + F_{\text{weighted\_conv}}.
\label{eq:coswin_msa}
\end{equation}

Similarly, for convolution fused shifted window attention, the original windows of the images are shifted by a stride of $\frac{M}{2}$. Afterward, the attention metric, i.e., CoSwin-Shifted-MSA(X), can computed as follows:
\begin{equation}
\text{CoSwin-Shifted-MSA}(X) = \left( \bigcup_{i=1}^{S} \text{MSA}(\text{Shift}(X_i)) \right) + F_{\text{weighted\_conv}}.
\label{eq:coswin_shifted_msa}
\end{equation}
\\
\noindent \textbf{Block Formulation:}  
Each CoSwin Transformer block consists of two consecutive attention-MLP sequences: one based on regular window attention and the other based on shifted window attention. Formally, two successive CoSwin blocks are computed as:
\begin{align}
\hat{\mathbf{z}}_l &= \text{CoSwin-MSA}(\text{LN}(\mathbf{z}_{l-1})) + \mathbf{z}_{l-1}, \nonumber \\
\mathbf{z}_l &= \text{MLP}(\text{LN}(\hat{\mathbf{z}}_l)) + \hat{\mathbf{z}}_l, \nonumber \\
\hat{\mathbf{z}}_{l+1} &= \text{CoSwin-Shifted-MSA}(\text{LN}(\mathbf{z}_l)) + \mathbf{z}_l, \nonumber \\
\mathbf{z}_{l+1} &= \text{MLP}(\text{LN}(\hat{\mathbf{z}}_{l+1})) + \hat{\mathbf{z}}_{l+1},
\end{align}
where $\mathbf{z}_l$ and $\hat{\mathbf{z}}_l$ represent the outputs of the convolution-fused MSA modules and MLPs at layer $l$, respectively.

\begingroup
\begin{table*}[t]
\centering
\renewcommand{\arraystretch}{1.16} % 
\caption{Top-1 validation accuracy comparison of proposed model with various state-of-the-art models when training from scratch.}
\resizebox{\textwidth}{!}{
\begin{tabular}{l|lccccc }
\toprule
\textbf{Method} & \textbf{Architecture}  & \textbf{CIFAR-10} & \textbf{CIFAR-100} & \textbf{MNIST} & \textbf{SVHN} & \textbf{T-ImageNet} \\
\midrule
ResNet18~\cite{cite_resnet} & Convolution  & 90.27\% & 66.46\% & 99.20\% & 97.76\% & 53.52\% \\ 
\hline
ResNet56~\cite{cite_resnet} & Convolution  & 94.63\% & 74.81\% & 99.21\% & 97.66\% & 56.5\% \\ 
\hline
MobileNetV2/1.4\cite{cite_mobilenet2} &  Convolution  & 92.81\% & 67.6\% & 99.41\% & 97.73\% & 58.4\% \\
\hline
ViT~\cite{cite_vit} & Transformer& 93.44\% & 72.75\% & 99.53\% & 97.79\% & 55.28\%\\
\hline
DeiT-Tiny~\cite{cite_deit} & Transformer  & 94.01\% & 71.40\% & 99.53\% & 97.73\% & 50.01\%\\ 
\hline
T2T~\cite{cite_t2t} & Transformer  & 95.24\% & 76.08\% & 99.54\% & 97.87\% & 58.54\%\\ 
\hline
SwinT~\cite{cite_swin} & Transformer  & 94.46\% & 76.72\% & 99.50\% & 97.81\% & 60.59\%\\ 
\hline
\textbf{CoSwin (Proposed)} & Transformer-Conv & \textbf{96.63\%} & \textbf{81.64\%} & \textbf{99.60\%} & \textbf{98.07\%} &\textbf{65.06\%}\\
\toprule
\end{tabular}
}

\label{table:comparision}
\end{table*}
\endgroup

\begin{figure*}[t]
    \centering
    \begin{minipage}[b]{0.335\textwidth}
        \centering
        \includegraphics[width=\textwidth]{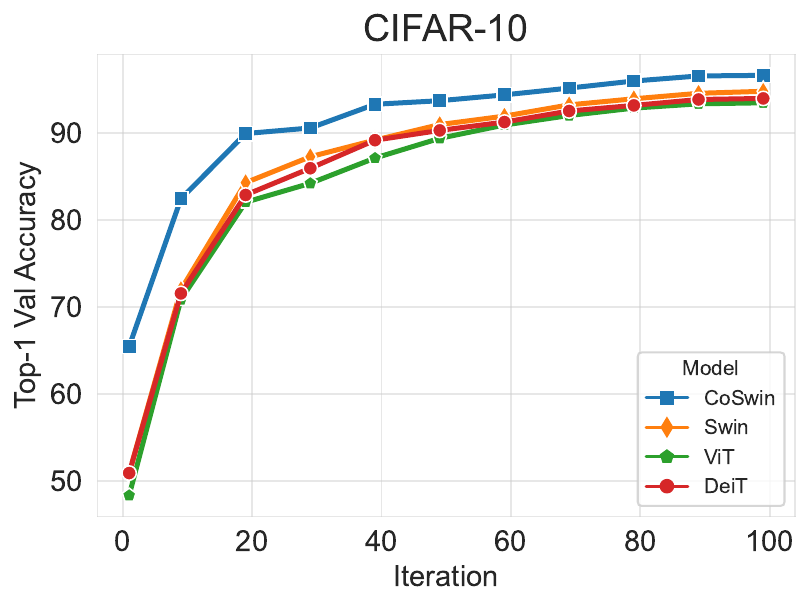}
    \end{minipage}%
    \begin{minipage}[b]{0.335\textwidth}
        \centering
        \includegraphics[width=\textwidth]{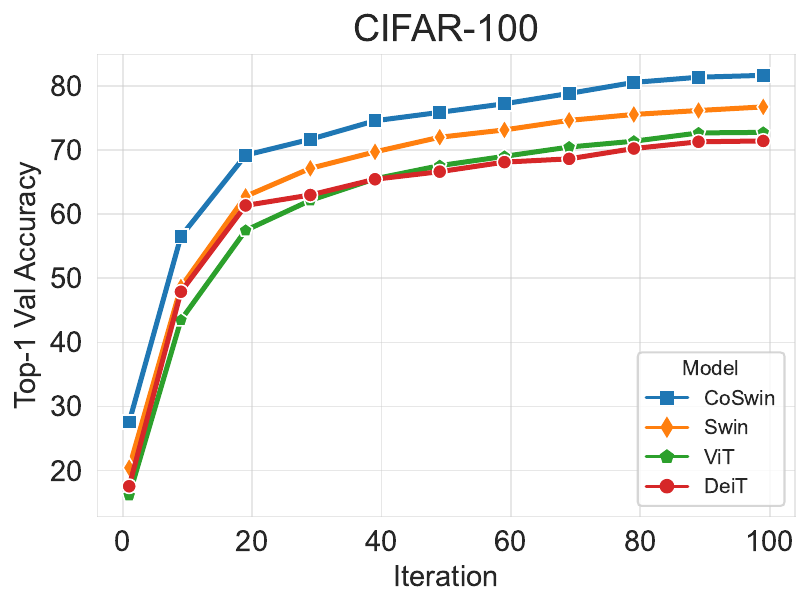}
    \end{minipage}%
    \begin{minipage}[b]{0.335\textwidth}
        \centering
        \includegraphics[width=\textwidth]{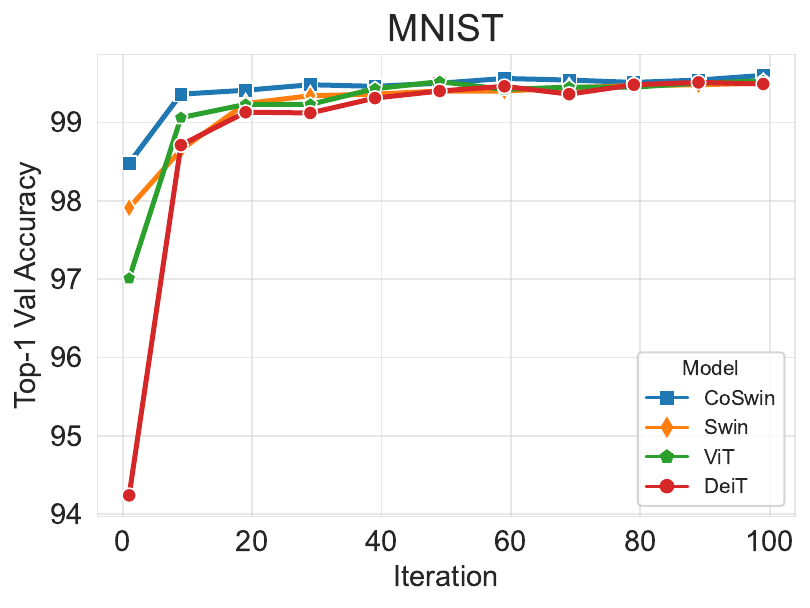}
    \end{minipage}
    
    \begin{minipage}[b]{0.335\textwidth}
        \centering       
        \includegraphics[width=\textwidth]{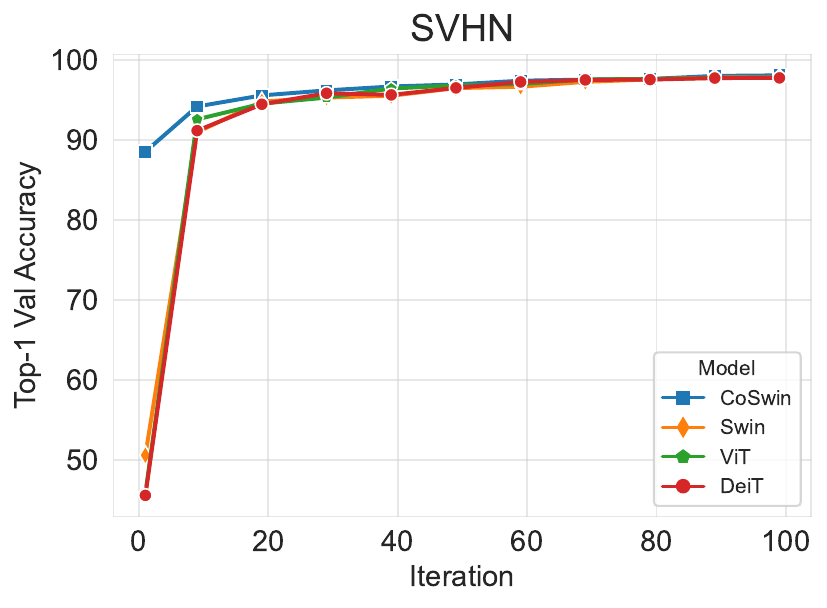}
    \end{minipage}%
    \begin{minipage}[b]{0.328\textwidth}
        \centering
        \includegraphics[width=\textwidth]{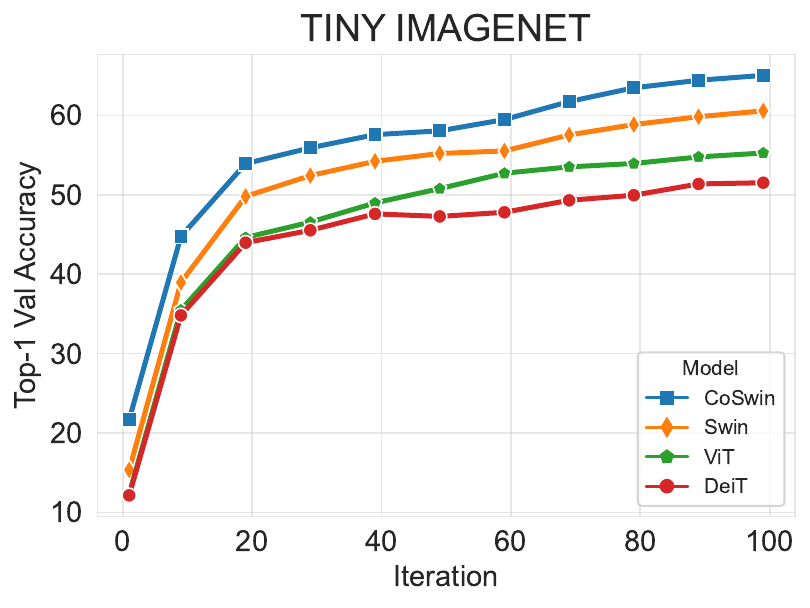}
    \end{minipage}%
    \begin{minipage}[b]{0.335\textwidth}
        \centering
        \hspace{1cm} 
    \end{minipage}
    \caption{Top-1 Accuracy Visualization over 100 epochs. (Top Left) on CIFAR-10 dataset. (Top Middle) on CIFAR-100 dataset. (Top Left) on MNIST dataset. (Bottom Left) on SVHN dataset. (Bottom Middle) on Tiny ImageNet dataset.}
    \label{fig:acc}
\end{figure*}

\section{Experiments}
In this section, we demonstrate the effectiveness of our proposed model. First, we describe our datasets and its composition. Next, we highlight our training specifications, including the parameters and experimental setup. We then compare the performance of the proposed model with that of other several state-of-the-art models. Lastly, we conduct an ablation study to further analyze the influence and contribution of various components of our model.

\subsection{Dataset}
We conduct our experiments on five benchmark datasets: CIFAR-10~\cite{cite_cifar}, CIFAR-100, MNIST~\cite{cite_mnist}, SVHN~\cite{cite_svhn} and Tiny Imagenet~\cite{cite_tiny_imagenet} (see Table ~\ref{table:dataset}), for the image classification task. 

\begin{table}[h]
\centering
\renewcommand{\arraystretch}{1.3}  
\caption{ overview of datasets.}
\resizebox{0.5\textwidth}{!}{
\begin{tabular}{c c c c c c }
\toprule
Dataset & CIFAR-10 & CIFAR-100 & MNIST & SVHN & T-Imagenet \\
\midrule
Size & $32\times32$ & $32\times32$ & $28\times28$ & $32\times32$ & $64\times64$  \\
Training Set & 50,000 & 50,000 & 60,000 & 73,257 & 100,000 \\
Validation Set & 10,000 & 10,000 & 10,000 & 26,032  & 10,000\\
Total Classes & 10 & 100 & 10 & 10 & 200\\
\hline
\end{tabular}
}
\label{table:dataset}
\end{table}
To maintain fairness in model comparison, we applied the same data augmentation techniques including filtering and transformation, consistently to each image in the dataset before feeding them into the proposed coswin and various baseline models.

\begin{figure*}[t]
\centering
\includegraphics[width=\textwidth]{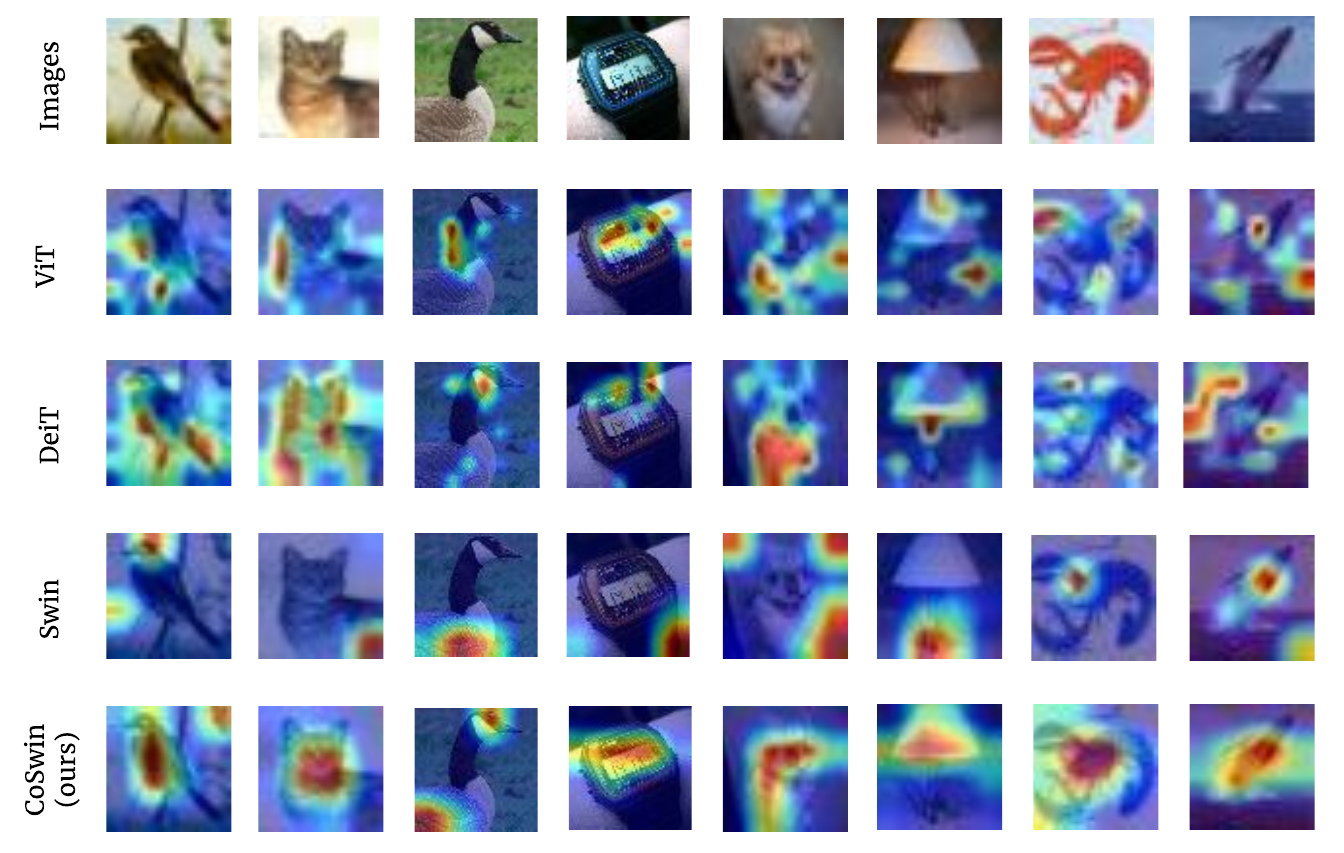}
\caption{Visualization of attention score using Grad-CAM.}
\label{fig:att_map}
\end{figure*}

\subsection{Training}
We implemented our proposed model using PyTorch~\cite{cite_pytorch} and trained it from scratch for 100 epochs on an NVIDIA V100 GPU. We applied standard data augmentation techniques such as MixUp~\cite{cite_mixup}, CutMix~\cite{cite_cut_mix}, Random Erasing~\cite{cite_random_erasing}, Repeated Augment~\cite{cite_repeat_augment}, and AutoAugment~\cite{cite_auto_augment}. In addition, Stochastic Depth~\cite{cite_stochastic_depth} was used as a regularization technique. The dataset was divided into batches of size 128, and training began with random parameter initialization. For optimization, we used AdamW~\cite{cite_adamw} with a weight decay of 0.05. We applied a stochastic drop rate of 0.1 and set the warm-up period to 10 epochs. Furthermore, we employed a learning rate scheduler to dynamically adjust the learning rate during training. The initial learning rate was set to 1e-3 and was gradually updated according to the scheduler.

\subsection{Comparison with the state of the art models}
Table~\ref{table:comparision} provides a comprehensive comparison of our proposed model against several state-of-the-art models. We selected both CNN-based architectures, such as ResNet18~\cite{cite_resnet}, ResNet56~\cite{cite_resnet}, and transformer-based models, including Vision Transformer (ViT)~\cite{cite_vit} and Swin Transformer~\cite{cite_swin}. Specifically, we used the small variant of each transformers and trained them from scratch. The default patch size was set to 4. For smaller images with resolutions below $32 \times 32$, the patch size was adjusted to 2. All models were trained under identical conditions to ensure a fair and consistent evaluation. Furthermore, Fig.~\ref{fig:acc} shows the top-1 validation accuracy curves of the Vision Transformer~\cite{cite_vit}, DeiT~\cite{cite_deit} and Swin Transformer~\cite{cite_swin} models in comparison to CoSwin over 100 epochs.

\subsubsection{Results on CIFAR-10/100}
We evaluated the models on the CIFAR-10 and CIFAR-100 datasets, both of which have a fixed resolution of $32 \times 32$. Training was conducted for 100 epochs, during which our model demonstrated outstanding performance. On the CIFAR-10 dataset, our proposed model achieved an accuracy of 96.63\%, surpassing all the competing models. Similarly, on the CIFAR-100 dataset, our model outperformed its counterparts by achieving an accuracy of 81.64\%, which indicates its ability to generalize effectively to more complex classification tasks. These results underscore the superiority of our model over existing CNN and transformer-based models.

\begingroup
\begin{table*}[h]
\centering
\caption{Ablation study of different components and performance evaluation.}
\renewcommand{\arraystretch}{2} 
\resizebox{\textwidth}{!}{ 
\begin{tabular}{cccc|ccccc }
\toprule
\textbf{Method} & \textbf{Convolution-1 }& \textbf{Convolution-2} & \textbf{Influence Weight} & \textbf{CIFAR-10} & \textbf{CIFAR-100} & \textbf{MNIST} & \textbf{SVHN} & \textbf{T-IMAGENET} \\
\midrule
a & \xmark & \xmark & \xmark &  \makecell{94.41\% \\ \textcolor{gray}{(-2.22)} }  & \makecell{76.72\% \\ \textcolor{gray}{(-4.92)} } & \makecell{99.50\% \\ \textcolor{gray}{(-0.10)}} & \makecell{97.81\% \\ \textcolor{gray}{(-0.26)}} & \makecell{60.59\% \\ \textcolor{gray}{(-4.47)}} \\
\hline
b & \cmark & \cmark & \xmark &  \makecell{95.93\%  \\ \textcolor{gray}{(-0.70)} } & \makecell{79.13\% \\ \textcolor{gray}{(-2.51)}} & \makecell{99.59\% \\ \textcolor{gray}{(-0.01)}} & \makecell{97.71\% \\ \textcolor{gray}{(-0.36)}} & \makecell{61.67\% \\ \textcolor{gray}{(-3.39)}} \\
\hline
c & \cmark & \xmark & \cmark & \makecell{ 96.17\% \\ \textcolor{gray}{(-0.46)}} & \makecell{80.67\% \\ \textcolor{gray}{(-0.97)}} & \makecell{ 99.57\% \\ \textcolor{gray}{(-0.03)}} & \makecell{97.99\% \\ \textcolor{gray}{(-0.08)}} & \makecell{64.77\% \\ \textcolor{gray}{(-0.029)}}\\
\hline
d & \cmark & \cmark & \cmark (Elementwise Add) & \textbf{96.63\%} &\textbf{ 81.64\%} & \textbf{99.60\%} & \textbf{98.07\%} & \textbf{65.06\%}
\\ 
\bottomrule
\end{tabular}
}
\label{table:ablation}
\end{table*}
\endgroup

\subsubsection{Results on MNIST}
To evaluate the performance of our model on simpler datasets, we conducted experiments on the MNIST dataset~\cite{cite_mnist}, which consists of gray-scale images with a resolution of 28 × 28. For this dataset, the patch size was set to 2 due to the lower image resolution. All models were trained from scratch under identical conditions. Our model performed well on this dataset compared to existing state-of-the-art methods, achieving an accuracy of 99.60\%. These results validate the versatility and efficiency of our approach across datasets with varying levels of complexity.

\subsubsection{Results on SVHN}
We also evaluated CoSwin's performance on the SVHN dataset which consists of 10 classes with images of size 32x32 pixels. Due to the relatively low image resolution, we set the default patch size to 2. Our model achieved 98.07\% top-1 validation accuracy surpassing existing state-of-the-art models. For reference, Swin Transformer and Vision Transformer, when adapted to the SVHN dataset, reached 97.81\% and 97.79\% accuracy respectively. These results shows our model's ability to effectively capture fine-grained features in the SVHN dataset.

\subsubsection{Results on Tiny ImageNet}
Finally, we assessed our model on the Tiny ImageNet dataset, a challenging subset of the large-scale ImageNet dataset. This dataset features images with 200 classes, making it an ideal benchmark for evaluating model performance. For these experiments, the patch size was set to default 4. Our model achieved a top-1 validation accuracy of 65.06\%, which represents an improvement of approximately 4.47 percentage points over the Swin Transformer under comparable architecture constraints. Additionally, our model outperformed the vision transformer and CNN-based models, such as ResNet variants, by a substantial margin.

The results from our experiments demonstrate the efficiency of our model in comparison with the leading state-of-the-art methods for small dataset challenges. These findings highlight the strengths of our model’s architecture in terms of ability to handle limited data, and strong generalization across diverse image classification tasks.

\subsection{Attention Visualization}
Fig.~\ref{fig:att_map} shows the class activation maps of the classified image as generated by the ViT (second row), DeiT (third row), Swin Transformer (fourth row), and our CoSwin model (fifth row), respectively. We visualize the attention map obtained at the final normalization layer of the transformer using Grad-CAM~\cite{cite_grad_cam}. From the obtained spatial heat map, it is evident that our model achieves a sharper focus on the main subject of the image than that in the existing transformer models. This enhanced attention is primarily due to the integration of the weighted convolution score within the transformer, which effectively captures both local features and global contextual information, including major inductive biases and makes it well-suited for visual analysis tasks with small datasets.

\subsection{Ablation Studies}
To study the influence of the fused convolution module and other components on the proposed model, we conduct ablation studies with default settings as shown in Table~\ref{table:ablation}. This experiment is performed over 100 epochs using a batch of size 128.

\subsubsection{Effect of convolution features in the attention module}
We remove the fused convolutional layers from the transformer block. Removing just the second convolution layer results in a 0.46\% drop in the Top-1 accuracy of CIFAR-10, 0.97\% in CIFAR-100, 0.03\% in MNIST, 0.08\% in SVHN, and 0.029\% in Tiny ImageNet, as shown in Table~\ref{table:ablation}. Furthermore, when the entire convolutional layers is removed, the Top-1 accuracy drops increase to 2.22\% in CIFAR-10, 4.92\% in CIFAR-100, 0.10\% in MNIST, 0.26\% in SVHN, and 4.47\% in Tiny ImageNet. This shows that self-attention alone does not allow the model to perform optimally in small-scale vision. However, when convolution features are applied to the attention mechanism, they complement each other by capturing local-global features and strong inductive biases which gradually improves generalization on small-scale vision tasks.

\subsubsection{Effect of weighted value in convolution fusion}
We study whether a weighted value during feature fusion is needed in our model. To investigate this, we add the convolution features directly to the attention without assigning a trainable weight. In this case, the model becomes more prone to overfitting and leads to 0.70\% accuracy drop in CIFAR-10. Similarly, accuracy drops 2.51\% in CIFAR-100, 0.01\% in MNIST, 0.36\% in SVHN, and 3.39\% in Tiny ImageNet, as shown in Table~\ref{table:ablation}b. These results show that removing the weight from the convolution layer causes the model to treat both convolution and attention score equally. However, in small scale datasets, depending on the image variations, there are cases where it is necessary to place more emphasis on the global context rather than the local.

\section{Conclusion}
In this paper, we presented CoSwin, a hybrid vision transformer architecture that integrates learnable local feature augmentation into the hierarchical shifted window attention mechanism. Motivated by the limitations of pure attention-based models in small-scale and data-constrained scenarios, CoSwin introduces a lightweight convolutional module to enrich spatial locality and reduce reliance on large-scale data and global attention alone. This local-global fusion is achieved through a dynamic weighting mechanism that adaptively balances convolutional features and multi-head shifted window attention outputs, enabling the model to capture both fine-grained spatial details and long-range semantic dependencies.
Our architecture retains the efficiency and scalability of window-based transformers while restoring critical inductive biases such as locality and translation equivariance properties essential for generalization on small-resolution or low-data regimes. Through extensive experiments on benchmark datasets including CIFAR-10, CIFAR-100, MNIST, SVHN, and Tiny ImageNet, CoSwin demonstrates consistent and significant performance gains over state-of-the-art CNN and transformer baselines. 
Future work may extend the architecture to dense prediction tasks such as semantic segmentation and object detection, or to domain-specific applications such as medical imaging, where it is especially important to balance fine local details with a broader global context.

\section*{Acknowledgment}
This work was supported in part by the National Science Foundation under Award \href{https://www.nsf.gov/awardsearch/showAward?AWD_ID=2346643}{\#2346643}.

\appendix

\section*{Theoretical Analysis: Convolution-Attention Fusion in Small-Scale Vision}
\label{sec:theory_smallscale}

In what follows, we provide a formal theoretical analysis on why the inclusion of the convolution feature map in the shifted attention mechanism of the proposed CoSwin architecture leads to improved generalization, reduced variance,
and efficient function approximation in small-scale vision tasks.

Let \(\mathcal{X} \in \mathbb{R}^{H \times W \times C}\) denote the input space of images, and \(\mathcal{Y} = \{1, \dots, K\}\) be the label space for \(K\) classes. Similarly, \(f_\theta: \mathcal{X} \rightarrow \mathbb{R}^K\) be a parameterized model with \(\theta\) parameter and \(L(f_\theta(x), y)\) denotes the cross-entropy loss. We study three hypothesis classes.

\subsection{Hypothesis Classes and Inductive Bias}

\smallskip

\noindent {\em Definition 1. Convolutional Hypothesis Space:} A function class 
\(\mathcal{H}_{\text{Conv}}\) contains all functions representable by convolutional layers with bounded kernel size \(k \times k\), weight sharing, and local connectivity:
\[
\mathcal{H}_{\text{Conv}} = \{ f(x) = \phi(W * x + b) \mid W \in \mathbb{R}^{k \times k \times C}, \phi \text{ is ReLU} \}.
\]

\smallskip

\noindent {\em Definition 2. Self-Attention Hypothesis Space:} A attention head over \(N\) image patches of dimension \(d\) is defined as:
\[
\mathcal{H}_{\text{Att}} = \text{Softmax}\left(\frac{QK^\top}{\sqrt{d}} + B\right)V,
\]
where \(Q, K, V \in \mathbb{R}^{N \times d}\) are the learned projections of the input and \(B\) denotes the relative positional bias.

\smallskip

\noindent {\em Definition 3. Fused Hypothesis Space:} Coswin augments the locally enhanced features from $\mathcal{H}_{\text{Conv}}$ with the global features extracted from $\mathcal{H}_{\text{Att}}$ through a shifted window using a learnable scalar fusion parameter \(\Gamma\):
\begin{align}
\mathcal{H}_{\text{CoSwin}} = \left( \bigcup_{i=1}^{S} \text{MSA$_{sw}$}(X_{i}) \right) + F_{\text{weighted\_conv}} 
\label{eq:hypo}
\end{align}
where \text{MSA$_{sw}$} represents shifted window multi head self attention within i$^{th}$ window of an image $X$.

The inductive bias of CoSwin constrains the hypothesis space to functions which combine translation-equivariant local features with long-range global dependencies. This results in a smaller effective capacity and better sample efficiency on small-scale datasets.

\subsection{Efficient gradient propagation}
CoSwin maintain stable optimization through residual path, pre-normalization, and local feature fusion. The gradient with respect to the input $X$ in Eq.~\ref{eq:hypo} is given by:

\begin{equation}
\small
\frac{\partial \mathcal{L}}{\partial X} =
\frac{\partial \mathcal{L}}{\partial \text{MSA$_{sw}$}(X)} \cdot 
\frac{\partial \text{MSA$_{sw}$}(X)}{\partial X}
+ \Gamma \cdot 
\frac{\partial \mathcal{L}}{\partial F_{\text{conv}}} \cdot 
\frac{\partial F_{\text{conv}}}{\partial X}
\end{equation}

Since convolutional operators have well-conditioned, bounded gradients due to their local linear structure, the convolutional fusion branch regularizes the overall gradient magnitude and mitigates sharp fluctuations resulting from softmax-based attention scores. This improves stability and prevents training collapse.

\subsection{Bias-Variance Decomposition}
In small-scale vision, the generalization error can be decomposed as
\begin{equation}
\mathbb{E}\!\left[(f(x) - f^*(x))^2\right] \;=\; \text{Bias}^2 + \text{Variance} + \sigma^2,
\end{equation}
Here, shifted window attention reduces bias by capturing dependencies across windows, while convolution reduces variance by enforcing translation and locality equivariance. The learnable fusion weight $\Gamma$ adaptively balances the two. As a result, CoSwin mitigates variance while controlling bias, which improves generalization in small-scale vision.

\newcommand{\BIBdecl}{\setlength{\itemsep}{0.5 em}}
\linespread{1.06}\selectfont
\bibliographystyle{IEEEtran.bst}
\bibliography{bibliography}
\balance

\end{document}